# TSDM: Tracking by SiamRPN++ with a Depth-refiner and a Mask-generator

Pengyao Zhao, Quanli Liu*, Wei Wang and Qiang Guo
Key Laboratory of Intelligent Control and Optimization for Industrial Equipment of Ministry of Education
Dalian University of Technology
Dalian, China
liuql@dlut.edu.cn

*Abstract*—In a generic object tracking, depth (D) information provides informative cues for foreground-background separation and target bounding box regression. However, so far, few trackers have used depth information to play the important role aforementioned due to the lack of a suitable model. In this paper, a RGB-D tracker named TSDM is proposed, which is composed of a Mask-generator (M-g), SiamRPN++ and a Depth-refiner (D-r). The M-g generates the background masks, and updates them as the target 3D position changes. The D-r optimizes the target bounding box estimated by SiamRPN++, based on the spatial depth distribution difference between the target and the surrounding background. Extensive evaluation on the Princeton Tracking Benchmark and the Visual Object Tracking challenge shows that our tracker outperforms the state-of-the-art by a large margin while achieving 23 FPS. In addition, a light-weight variant can run at 31 FPS and thus it is practical for real world applications. Code and models of TSDM are available at https://github.com/lql-team/TSDM.

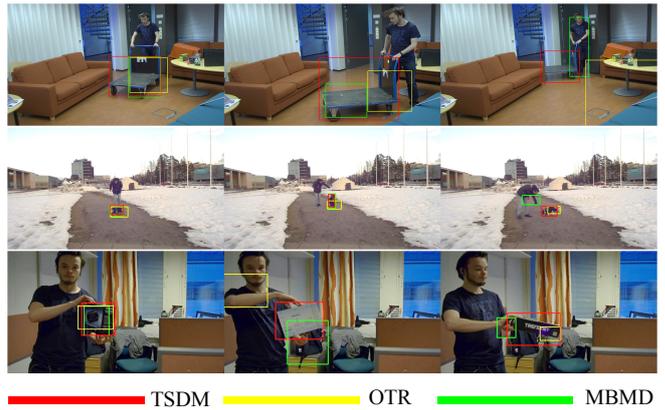

Fig. 1. Comparison examples of TSDM with two state-of-the-art trackers. The tracking targets from top to bottom are a cart, a dog and a box respectively. OTR [12], a RGB-D tracker, based on correlation filters ranked best in Princeton Tracking Benchmark [28]. MBMD [30] based on a deep network, with an online-update classifier, won the VOT-LT2018 [13] champion. TSDM is the proposed tracker.

## I. INTRODUCTION

Visual object tracking is a fundamental problem in computer vision tasks, such as autonomous driving, visual surveillance and human-computer interface. It is usually defined as the process that a tracker gets the target state (represented by a bounding box) in the first frame of the video, and then estimates the target state in subsequent frames. Nowadays, most popular methods for visual object tracking are almost always based on color information and rarely use depth information which is useful for improving tracking.

With the improvement of 3D acquisition technology, it is easy to obtain the depth image by a Lidar, a ToF camera or Binocular stereo vision technology. The depth image provides the distance from each pixel to the depth camera and reflects the edges of objects clearly. However, so far, depth information has contributed little to tracking. **The main obstacle** is that the tracker requires constant information (such as color), but the target depth distribution may change a lot when the target moves.

**Contribution**: (1) We propose **TSDM**, a new RGB-D tracker, which can ignore background distractors and output an accurate target state (see Fig. 1). (2) We first propose two novel depth modules (M-g and D-r) which can overcome the obstacle above and make use of depth information effectively. (3) We propose a M-g simulated data augmentation method. Benefiting from this method, SiamRPN++ [18] can be retrained to work better with the M-g module and achieve the further performance improvement.

The rest of the paper is organized as follows. Section II introduces some popular RGB trackers and representative RGB-D trackers; Section III describes the proposed tracker; Section IV evaluates the proposed tracker on two benchmarks and does an ablation study; Section V concludes the paper.

## II. RELATED WORK

### A. RGB tracking

**Correlation-based trackers** introduces circular correlation from the signal processing field into the visual tracking task and solves the target template matching problem by performing operations in frequency domain. Representative works include KCF [7], CSR-DCF [21] and ECO [4], etc. KCF trains correlation filters by kernel ridge regression, with the running speed above 150 FPS and only a few lines of core code. CSR-DCF constructs a spatial reliability map to constrain the correlation filter learning and uses Alternating

Direction Method of Multipliers to train the filters. ECO improves C-COT [5] (an excellent tracker using CNN feature, CN feature and HOG feature) from model size, training set size and model update. It became one of the most accurate trackers before 2017.

**End-to-end trackers** solve the target template matching problem through off-line training, such as Siamese series [1, 19, 18] and ATOM [3]. SiamFC builds an end-to-end tracking framework to estimate the region-wise feature similarity between two frames. Based on SiamFC, SiamRPN introduces the RPN module [24] to calculate the target boundary directly instead of multi-scale testing. SiamRPN++ applies a deeper network (Resnet [8]) for further performance jump. ATOM differs from the Siamese series. It calculates the target boundary by extensive off-line training, but locates the target through an online-update CNN classifier. So far, extensions of Siamese series and ATOM have shown amazing performance better than correlation-based trackers in VOT2019 [14].

### B. RGB-D tracking

So far, the number of RGB-D trackers is much less than RGB trackers. Some representative works are as follows.

**DS-KCF** [6]**:** It was proposed by Hannuna et al., with KCF as the core. It uses KCF to estimate the target bounding box first and then segments the target finely based on the depth information. The segmentation result is used for the target re-estimation and occlusion judgment. In order to obtain this accurate target segmentation above, DS-KCF constructs a depth information histogram and applies a clustering scheme to find out the distribution of minimum depth cluster. Then the connected component of this depth distribution on the image plane is taken as the target segmentation.

**3D-T** [2]**:** It was proposed by Bibi et al., with particle filter framework as the core. When the "depth-normalized" size of the target decreases below the set threshold, the tracker considers that the target is occluded. More surprising than building a 3D tracker, the paper synchronizes color information and depth information based on optical flow and the interpolation operation. This reduces the color-depth mismatch to some extent.

**Ca3dms** [16]**:** It was proposed by Y. Liu et al., which extends the traditional 2D mean-shift method to 3D with some mechanisms to further boost the robustness. It sets up a target sphere area to contain the target and a bounding sphere area to contain the surrounding background. If the number of particles in both the two area changes little, Ca3dms updates the template color histogram to match the target. The authors also proposed a strategy to overcome short-term total occlusion. When the target is completely occluded, the number of particles in the target sphere area falls to a very low level. Then the tracker tracks the occluder until the target is visible.

**DM-DCF** [10] **& OTR** [12]**:** They were proposed by Cart in 2018 and 2019 respectively and use CSR-DCF [21] as the core. DM-DCF uses two constantly updated gaussian functions to fit the target depth distribution and the background depth distribution respectively, and produces a foreground probability image through these two depth distributions. Then, it segments the foreground probability image by OTSU [23] for generating the spatial reliability map of CSR-DCF. Finally, it uses CSR-DCF to estimate the target bounding box. Another innovation is that the tracker judges target occlusion state through the filter response and the number of mask pixels in the bounding box together.

In 2019, the author proposed OTR based on DM-DCF. Prior to our method, OTR is the best-performing tracker in the Princeton Tracking Benchmark [28]. Compared with DM-DCF, OTR uses both of color and depth information to build the spatial reliability map of CSR-DCF. Moreover, as the aspect ratio of the target changes, it constructs multiple templates and trains multiple filters to overcome the effect of target 3D-rotation on tracking.

### C. Summary

Based on the above related works from 2013 to 2019, two conclusions can be drawn:

- Since 2019, the performance of end-to-end RGB trackers has been superior to correlation-based RGB trackers.
- Some successful RGB-D trackers select an excellent RGB tracker as the core, and use depth information as an auxiliary to improve the performance of the core.

Therefore, by taking SiamRPN++ as the core method to set a high lower-limit of the performance, two novel assistant modules are proposed for using depth information efficiently. Together, these three parts constitute **TSDM**, which performs better than SiamRPN++ on popular evaluation benchmarks.

### III. METHODOLOGY

### A. The tracking pipeline of TSDM

TSDM consists of a RGB tracker core and two assistant modules. The core is **SiamRPN++,** which takes an image pair $(Z, X)$ as input and outputs the target bounding box in the current frame. Where, $Z$ is the template image (contains the target appearance information) obtained from the first frame, and $X$ is the larger candidate search image in the current frame. The model of SiamRPN++ is described as Eq. 1.

$$f(Z, X) = \phi(Z) \star \phi(X) \quad (1)$$

A deep backbone, Resnet-50 is used as the feature extractor $\phi(\cdot)$, which yields two feature maps from $(Z, X)$. Where, $\star$ denotes the cross correlation layer which measures the region-wise feature similarity between $(Z, X)$, and outputs a confidence score map and a spatial adjustment map for each anchor. Finally the adjusted anchor with the highest confidence score is taken as the target bounding box (for more details please refer to [18]).

The two assistant modules in TSDM are **Mask-generator** and **Depth-refiner**, which change the input and the output of the core respectively. The tracking pipeline of TSDM is shown as Fig. 2, and it can be divided into three steps as follows.

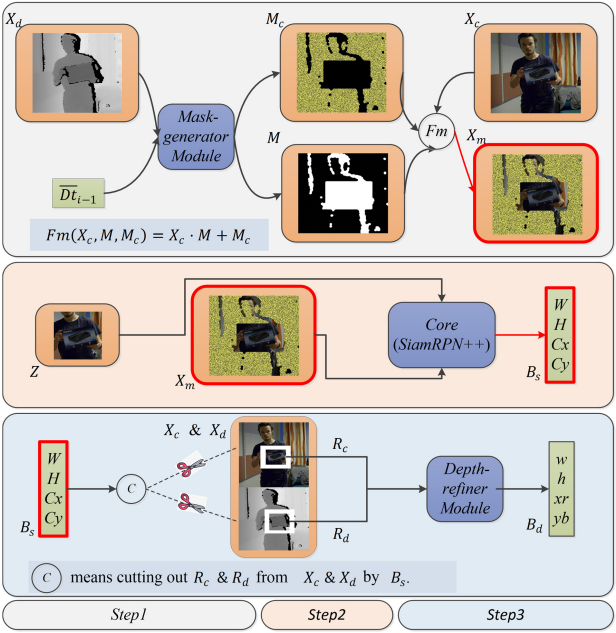

Fig. 2. The tracking pipeline of TSDM for the frame $i$. **In step1**, $X_d$ is the depth candidate search image obtained from the current depth frame; $X_c$ is the color candidate search image obtained from the current color frame; $\overline{Dt}_{i-1}$ is the average value of the target depth in the previous depth frame; $M$ and $M_c$ are background mask images generated by M-g; $X_m$ is $X_c$ with the background masks. **In step2**, $Z$ is the color template image obtained from the first color frame; $B_s$ is the target bounding box estimated by the core. **In step3**, $R_c$ and $R_d$ are the input of D-r; $B_d$ is the refined target bounding box computed by D-r.

- Step 1: Input $X_d$ and $\overline{Dt}_{i-1}$ into M-g to get $M$ and $M_c$. Then use $F_m(\cdot)$ to get $X_m$.
- Step 2: Input $Z$ and $X_m$ into the core. Then the core outputs the target bounding box $B_s$.
- Step 3: Cut out $R_c$ and $R_d$ from $X_c$ and $X_d$ by $B_s$ respectively. Then input $R_c$ and $R_d$ into D-r to get the refined target bounding box $B_d$.

*B. Mask-generator module*

In tracking videos, there are usual background distractors similar to the template image. Once the target appearance changes, the target score (Here, the **object score** means the **confidence score** of the anchor nearest to the **object** in SiamRPN++) will decrease. At this time, the distractor score may exceed the target, resulting in a tracking drift. M-g has two functions: (1) reduce the interference of background distractors. (2) clear out some image information irrelevant to the target in the current frame, which reduces difficulty of the target template matching. Concretely speaking, for achieving these two functions above, M-g generates two background mask images ($M$ and $M_c$) by Eq. 2 and Eq. 3 respectively.

$$M(x,y) = \begin{cases} 1, & \overline{Dt}_{i-1}/2 < D_i(x,y) < 2\overline{Dt}_{i-1} \\ 1, & \begin{cases} |x - C_{x_{i-1}}| < 0.75 w_{i-1} \\ |y - C_{y_{i-1}}| < 0.75 h_{i-1} \end{cases} \\ 0, & \text{otherwise} \end{cases} \quad (2)$$

$M$ is a 2-value image for clearing out the background of $X_c$. $\overline{Dt}_{i-1}$ is obtained accurately by using Otsu [23] in the previous estimated bounding box. $D_i(x,y)$ represents the depth value at every position of the current depth frame. $(C_{x_{i-1}}, C_{y_{i-1}})$ and $(w_{i-1}, h_{i-1})$ are the center position and size of the target in the previous frame respectively. Obviously, the useful pixels in $X_c$ within the depth range are kept. However, to be on the safe side, the pixels close to the previous estimated bounding box are also kept (overcome the color-depth mismatch).

$$M_c(x,y,c) = \rho(x,y,c) \times [1 - M(x,y)], \ c \in \{1,2,3\} \quad (3)$$

$M_c$ is a color image for coloring the background area of $X_c$. $\rho(x,y,c)$ represents the color distribution of $M_c$. As a spatial constraint, $[1 - M(x,y)]$ constrains the color area of $M_c$.

**$M_c$ color selection:** In essentially, $M_c$ enhances the target-background difference to make the target template matching easier (see Fig. 2–*step2*). So two colors are selected for $M_c$, which contrast sharply with the target average color in HSV domain. Specifically, Hue (H) difference of the three colors (the two colors for $M_c$ and the average target color) to one another is 120 degrees. Saturation (S) and Value (V) of the two colors are set to 100% and 70% respectively. These two colors are randomly distributed in $M_c$ (see Fig. 2–*step1*). Section IV-C shows that selecting two colors for $M_c$ is better than one.

**M-g stop-restart strategy:** M-g should automatically stop to avoid masking the real target when a transient tracking drift happens. Therefore, the elaborate strategy is proposed:

- If the value of $|\overline{Dt}_i - \overline{Dt}_{i-1}|$ is larger than $\gamma$ (a small positive constant) and the target score is lower than $\mu_1$, M-g will stop.
- If the target score is lower than $\mu_2$ ($\mu_2 < \mu_1$), M-g will also stop.
- If the target score is higher than $\mu_3$ (a convincing level), M-g will restart.

This strategy allows M-g to keep working when the target score is not very low and the target depth changes smoothly, but stops M-g working when the target score is very low. Therefore, this strategy strikes a balance between the effect and risk of M-g use.

**M-g simulated data augmentation:** Since the input of the core is $(Z, X_m)$ instead of $(Z, X_c)$, the core needs to be retrained. However, the lack of the suitable large-scale RGB-D datasets makes it difficult to use M-g to generate enough training samples ($X_m$). To simulate M-g, a M-g simulated data augmentation method is specially proposed for RGB datasets. Concretely speaking, first the two mask colors for $X_m$ are calculated based on the target color (like **$M_c$ Color selection**), then several rectangular color masks are created with random different sizes and aspect ratios, and finally these color masks are randomly placed outside the target bounding box. Section IV-C shows that retraining the core with the augmented data can improve the target locating accuracy.

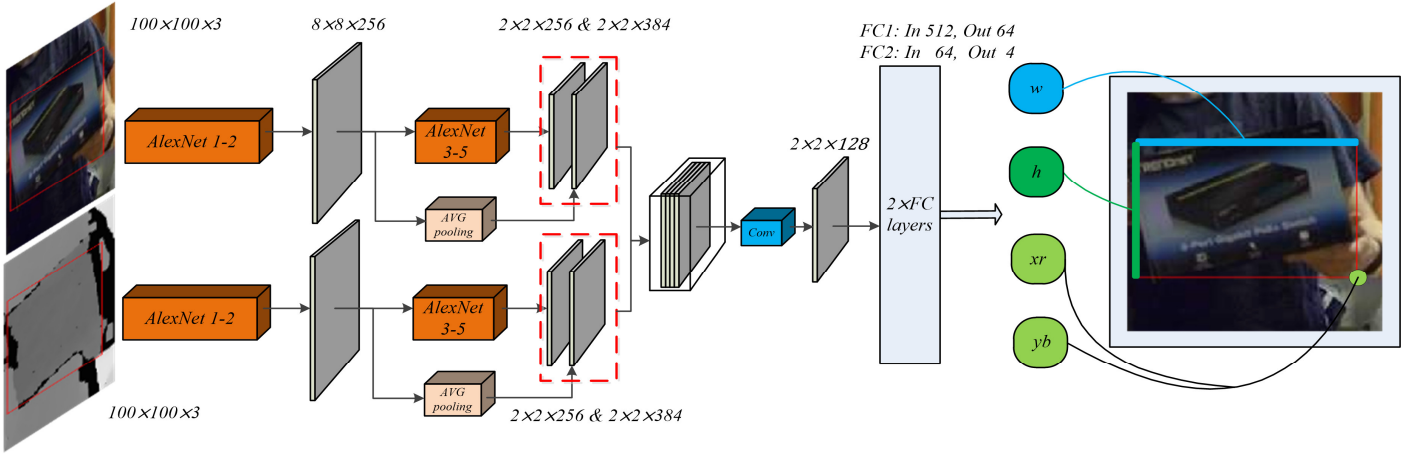

Fig. 3. Full architecture of the information fuse network. The color and depth images are first adjusted to the fixed size (100×100×3). Alexnet (layer 1-5) is used as the backbone, where the output of layer 5 and layer 2 (through the average pooling) are combined for semantic and spatial feature. Features from the two branches fuse together through one 1×1 convolution layer (blue Conv). A two-layer fully connected net regresses the refined target bounding box $(w, h, xr, yb)$.

## C. Depth-refiner module

**Fundamental assumption:** The core of our tracker only uses color information to estimate the target state. However the state can be refined by depth information which usually reflects the object outline directly. So Depth-refiner is proposed to optimize the target bounding box. It is based on a fundamental assumption that **the bounding box estimated by the core contains the whole target.** The assumption can be described as follows.

$$R_{gt} \subseteq R_{core} \quad (4)$$

Where, $R_{core}$ is a rectangular area (a pixel set) in the target bounding box estimated by SiamRPN++, and $R_{gt}$ is the minimum rectangular area (a pixel set) which contains the target. Based on this assumption, D-r can easily improve the tracker performance just by cutting out no-target area from $R_{core}$, and give a smaller and more precise target bounding box.

**Pretreatment:** How can we make sure that $R_{core}$ contains the whole target? Considering the characteristics of the core, the following two ways are adopted:

- Use Non-Maximum Suppression (NMS) with a high suppressed threshold $\alpha_1$ to merge multiple anchors into a larger region instead of the highest score anchor. This way can make $R_{core}$ larger reasonably.
- Set a scale amplification factor $\alpha_2$ ($\alpha_2 > 1$) to make $R_{core}$ larger.

**Information Fusion Network:** In essentially, D-r can be treated as an information fusion network. It uses depth information to optimize the target state, and meanwhile, color as the correction information overcomes the slight color-depth mismatch. By integrating the two information, the network can output a more accurate bounding box than $R_{core}$. The full architecture of D-r is shown in Fig. 3.

In the network, we adopt double Alexnet [15] rather than a single new 4-channel-input net as the backbone for a quick transfer learning; the 1×1 convolution layer fuses cross-channel information and reduces feature map dimension; the two fully connected layers are used to output the refined bounding box.

**Training Details:** The SUNRGBD [27] is used as our training dataset, which is composed of NYU-depth-v2 [26], Berkeley-B3DO [9] and SUN3D [29]. It contains 10000 RGB-D images and 3D bounding box annotations for objects in each image.

The larger region (target with surrounding background) is fed into the network as Fig. 3, and the network is trained to output the target bounding box size $(w, h)$ and the bottom-right corner position $(xr, yb)$. Then SmoothL1 function in PyTorch framework is used to measure the error between the predictions and the ground truth. SmoothL1 is defined as Eq. 5.

$$Smooth_{L1}(x) = \begin{cases} 0.5x^2, & |x| < 1 \\ |x| - 0.5, & \text{otherwise} \end{cases} \quad (5)$$

The complete loss function $\mathcal{L}$ is defined as Eq. 6, where $(\cdot)_{gt}$ represents the ground truth.

$$\mathcal{L} = Smooth_{L1}\left(\frac{w-w_{gt}}{w_{gt}}\right) + Smooth_{L1}\left(\frac{h-h_{gt}}{h_{gt}}\right) \\ + Smooth_{L1}\left(\frac{xr-xr_{gt}}{xr_{gt}}\right) + Smooth_{L1}\left(\frac{yb-yb_{gt}}{yb_{gt}}\right) \quad (6)$$

The Mini-Batch Gradient Descent (MBGD) is used to train the model with setting the mini-batch size as 16. The learning rate is linearly decreased from 0.05 to 0.0001 for whole 20 epochs. The backbone isn't trained in the first 3 epochs.

## I. EXPERIMENT

### A. Benchmarks & experiment Details

The proposed method is evaluated on two popular benchmarks, including the Princeton tracking benchmark (PTB) [28] and Visual Object Tracking challenge (VOT) [14] as follows. (1) PTB was presented by Song et al. in 2013, which was the largest RGB-D benchmark at that time. It includes various challenging categories such as Occlusion, Fast, Rigid, Animal etc., but no outdoor scenes. Moreover, videos in PTB are usually short. (2) VOT has been successfully held for 7 sessions in the past few years and released a RGB-D benchmark as its sub-project, called VOT-RGBD2019. VOT-RGBD2019 includes outdoor scenes, and many long videos. However, its baseline scale and video number are smaller. Therefore, the complementary advantages between the two benchmarks (Details see Table I) make it reasonable to use them for evaluating the proposed tracker together.

TABLE I. COMPARISION BETWEEN PTB AND VOT-RGBD2019

| Index | PTB | VOT-RGBD2019 |
|---|---|---|
| Number of videos | 95 ✓ | 80 |
| Average length of videos | 214 | 1274 ✓ |
| Ratio of Outdoor scenes | 0% | 12.5% ✓ |
| Number of baseline trackers | 38 ✓ | 12 |

All experiments are run on a single desktop computer with Intel Core i7-3.00GHz and a NVIDIA Titan X PASCAL GPU. In addition, set $\gamma = 0.01 \times Max(D_i(x,y))$, $\mu_1 = 0.65$, $\mu_2 = 0.55$ and $\mu_3 = 0.92$ to make a specific M-g stop-restart strategy (in Section III-B). Set $\alpha_1 = 0.7$ and $\alpha_2 = 0.1$ to increase $R_{core}$ size (in Section III-C Pretreatment).

### A. Performance on VOT-RGBD2019 & PTB

On the VOT-RGBD2019, the TSDM is compared with published baseline trackers provided, where OTR [12], CSR-depth [11], and Ca3dms [16] are 3D models, and other trackers [30, 22, 20] do not use depth information but perform well in the long-term tracking. As shown in the Fig. 4, TSDM has the highest AUC score (0.5351) and it is worth mentioning that the performance of TSDM is much better than other trackers in the **Threshold range** of 0.2-0.8. This means that once our tracker catches the target, the estimated bounding box will be very close to the ground truth.

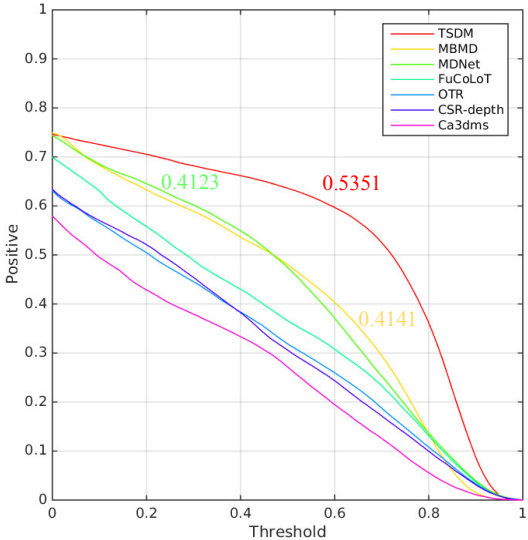

Fig. 4. Comparison of TSDM and the baseline in VOT-RGBD2019. The evaluation result is presented by different color lines, calculated as the ratio of frames with an intersection over union (IOU) overlap exceeding a threshold. Trackers are ranked through the area-under-the-curve (AUC) score.

On the PTB, TSDM is compared with 7 trackers, including the top 5 trackers [12, 17, 2, 11, 16] in the baseline and 2 trackers [10, 6] introduced in the Section II. All these trackers use depth information but are not based end-to-end CNN. Table II shows the average IOU overlap of each category and the overall average IOU overlap. Our tracker performs the best for **Overall**. Though our tracker wins first place in 7 categories, but performs poorly in **Human** and **Occ.** The reason is that videos of these two categories show target occlusion multiple times, but our method focuses on tracking accuracy, not dealing with occlusion as other trackers.

### B. Further analysis

Theoretically speaking, a double-color background mask is more readily identifiable than a single-color, and the retrained core can learn to use the surrounding mask to locate the target. The three cases are evaluated through 20 long videos in VOT-RGBD2019, which are separately Single-color Mask, Double-color Mask and Double-color Mask + Core Retraining (CR). The result proves that the double-color background mask achieves a 4.4% gain in terms of average IOU to the single-color, and retraining the core achieves a 5.8% relative gain to not retraining. A typical example is shown as Fig. 5.

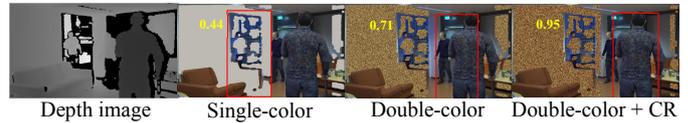

Depth image  Single-color  Double-color  Double-color + CR

Fig. 5. A typical comparison example of the three cases including Single-color, Double-color and Double-color + CR. The top-left yellow numbers and the red rectangles represent the target score and bounding box respectively (estimated by the core).

### C. Ablation study & Backbone degradation

On VOT-RGBD2019, the performance of SiamRPN++, TSDM and some variants are compared. Fig. 6 shows that adding both M-g and D-r to SiamRPN++ improves the accuracy by almost 10%. In addition, all variants still almost run at the real-time speed.

M-g can decrease the tracking difficulty significantly (see Fig. 2–step2). Therefore we try giving the core a light-weight backbone in order to increase the running speed when M-g is working. As shown in Fig. 6, MobileNetv2 [25] replaces Resnet-50 to increase the running speed to 31 FPS with only 2.6% reduction of the accuracy.

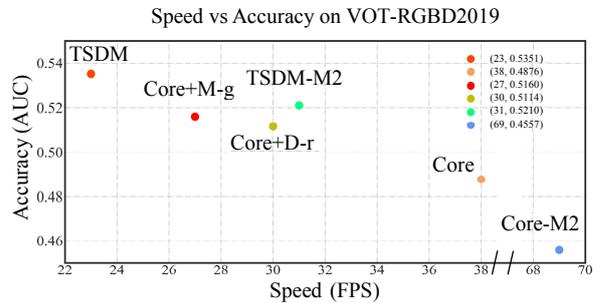

Fig. 6. Speed and Accuracy of TSDM, core and their variants on VOT-RGBD2019. M2 represents that the backbone of the core is MobileNetv2 instead of Resnet-50.

TABLE II.  COMPARISION OF TSDM AND THE BASELINE ON PTB (USE THE PTB PROTOCOL).

| Mehtod | Average IOU overlap | | | | | | | | | | | |
|---|---|---|---|---|---|---|---|---|---|---|---|---|
| | Overall | Human | Animal | Rigid | Large | Small | Slow | Fast | Occ. | No-Occ. | Passive | Active |
| TSDM | **0.792** | 0.71(6) | **0.85(1)** | **0.86(1)** | 0.77(2) | **0.81(1)** | **0.87(1)** | **0.76(1)** | 0.69(5) | **0.94(1)** | 0.84(3) | **0.78(1)** |
| OTR [12] | 0.769 | 0.77(2) | 0.68(3) | 0.81(3) | 0.76(4) | 0.77(2) | 0.81(2) | 0.75(2) | 0.71(2) | 0.85(4) | **0.85(1)** | 0.74(2) |
| ECO-TA [17] | 0.754 | 0.77(3) | 0.65(5) | 0.80(4) | 0.77(3) | 0.74(4) | 0.79(5) | 0.41(8) | 0.68(6) | 0.85(3) | 0.84(2) | 0.72(4) |
| 3D-T [2] | 0.750 | **0.81(1)** | 0.64(6) | 0.73(8) | **0.80(1)** | 0.71(7) | 0.75(8) | 0.75(3) | **0.73(1)** | 0.78(6) | 0.79(7) | 0.74(3) |
| CSR-rgbd++ [11] | 0.740 | 0.77(4) | 0.65(4) | 0.76(7) | 0.75(5) | 0.73(5) | 0.80(4) | 0.72(4) | 0.70(3) | 0.79(5) | 0.79(6) | 0.72(5) |
| Ca3dms [16] | 0.737 | 0.66(8) | 0.74(2) | 0.82(2) | 0.73(6) | 0.74(3) | 0.80(3) | 0.71(6) | 0.63(8) | 0.88(2) | 0.83(4) | 0.70(6) |
| DM-DCF [10] | 0.726 | 0.76(5) | 0.58(8) | 0.77(5) | 0.72(7) | 0.73(6) | 0.75(7) | 0.72(5) | 0.69(4) | 0.78(8) | 0.83(5) | 0.69(7) |
| DS-KCF [6] | 0.693 | 0.67(7) | 0.61(7) | 0.76(6) | 0.69(8) | 0.70(8) | 0.75(6) | 0.67(7) | 0.63(7) | 0.78(7) | 0.79(8) | 0.66(8) |

## V. CONCLUSION

A RGB-D tracking architecture, namely TSDM, is proposed with explicit modules for foreground-background separation and target state optimization. For foreground-background separation, a background mask scheme is applied to make the target template matching easier. The empirical evidence shows that the double-color background mask is better than the single-color. Considering the slight mismatch between color information and depth information, an information fusion network is designed to optimize the target bounding box. Comprehensive experiments on the two complementary RGB-D tracking benchmarks show that the proposed tracker can estimate more accuracy target state than the other state-of-the-art trackers and meet the real-time requirement. Moreover, our next work is to build a unified framework with the same characteristics as the proposed tracker.

ACKNOWLEDGMENT

This research was supported by the National Natural Science Foundation of China (61773085).